\documentclass{notices}
\usepackage{amsfonts,amssymb,amsmath,amscd,multirow}

\usepackage{comment}
\usepackage{flushend}

\usepackage{amsmath}
\usepackage{amsthm}
\usepackage{amsfonts}
\usepackage{dsfont}
\usepackage{graphicx}
\usepackage{soul}
\usepackage{rotating}
\usepackage{empheq}

\usepackage{hyperref}
\hypersetup{
    colorlinks=true,     
    urlcolor=magenta,
    }

\usepackage{xr}
\makeatletter
\newcommand*{\addFileDependency}[1]{
  \typeout{(#1)}
  \@addtofilelist{#1}
  \IfFileExists{#1}{}{\typeout{No file #1.}}
}
\makeatother

\usepackage{url}
\usepackage{comment}
\usepackage{float}
\usepackage{comment}
\usepackage[version=4]{mhchem}

\def \bb{b}
\def \bx{\boldsymbol{x}}
\def \bW{W}

\def \by{\boldsymbol{y}}
\def \bz{\boldsymbol{z}}

\def \bb{\boldsymbol{b}}
\def \bc{\boldsymbol{c}}

\def \bw{\boldsymbol{w}}

\def \bz{\boldsymbol{z}}
\def \bA{\boldsymbol{A}}

\def \bW{\boldsymbol{W}}

\def\R{{\mathbb R}}

\usepackage{xcolor}

\date{}

\title{\bf On the Geometry of Deep Learning}

\author{
  Randall Balestriero $^\diamond$
  \affil{Randall Balestriero is an Assistant Professor at Brown University. His email address is {\sf rbalestr@brown.edu}.
  }
  \and
  Ahmed Imtiaz Humayun $^\diamond$
  \affil{Ahmed Imtiaz Humayun is a PhD student at Rice University. His email address is {\sf imtiaz@rice.edu}.
   }
  \and
  Richard G.\ Baraniuk
  \affil{Richard Baraniuk is the C.\ S.\ Burrus Professor at Rice University. His email address is {\sf richb@rice.edu}.
  }
}

\begin{document}

\maketitle
\def\thefootnote{$^\diamond$}\footnotetext{Equal contributions}\def\thefootnote{\arabic{footnote}}

\subsection*{Introduction}

Machine learning has significantly advanced our ability to address a wide range of difficult computational problems and is the engine driving progress in modern artificial intelligence (AI).
Today's machine learning landscape is dominated by {\em deep (neural) networks}, which are compositions of a large number of simple parameterized linear and nonlinear operators.
An all-too-familiar story of the past decade is that of plugging a deep network into an engineering or scientific application as a black box, learning its parameter values using copious training data, and then significantly improving performance over classical task-specific approaches based on erudite practitioner expertise or mathematical elegance.

Despite this exciting empirical progress, however, the precise mechanisms by which deep learning works so well remain relatively poorly understood, adding an air of mystery to the entire field.
Ongoing attempts to build a rigorous mathematical framework have been stymied by the fact that, while deep networks are locally simple, they are globally complicated.
Hence, they have been studied primarily as ``black boxes'' and mainly empirically.
This approach greatly complicates analysis to understand both the success and failure modes of deep networks.
This approach also greatly complicates deep learning system design, which today proceeds alchemistically rather than from rigorous design principles.
And this approach greatly complicates addressing higher level issues like trustworthiness (can we trust a black box?), sustainability (ever-growing computations lead to a growing environmental footprint), and social responsibility (fairness, bias, and beyond).

In this paper, we overview one promising avenue of progress at the mathematical foundation of deep learning: the connection between deep networks and function approximation by {\em affine splines} (continuous piecewise linear functions in multiple dimensions).
In particular, we overview work over the past decade on understanding certain geometrical properties of a deep network's affine spline mapping, in particular how it tessellates its input space.
As we will see, the affine spline connection and geometrical viewpoint provide a powerful portal through which to view, analyze, and improve the inner workings of deep networks.

There are a host of interesting open mathematical problems in machine learning in general and deep learning in particular that are surprisingly accessible once one gets past the jargon.
Indeed, as we will see, the core ideas can be understood by anyone knowing some linear algebra and calculus.
Hence, we will pose numerous open questions as they arise in our exposition in the hopes that they entice more mathematicians to join the deep learning community.

The state-of-the-art in deep learning is a rapidly moving target, and so we focus on the bedrock of modern deep networks, so-called feedforward neural networks employing piecewise linear activation functions.
While our analysis does not fully cover some very recent methods, most notably transformer networks, the networks we study are employed therein as key building blocks.
Moreover, since we focus on the affine spline viewpoint, we will not have the opportunity to discuss other interesting geometric work in deep learning, including tropical geometry \cite{zhang2018tropical} and beyond.
Finally, to spin a consistent story line, we will focus primarily on work from our group; we will, however, review several key results developed by others.
Our bibliography is concise, and so we invite the interested reader to explore the extensive works cited in the papers we reference.

\subsection*{Deep Learning}

\noindent{\bf Machine learning in 200 words or less.} 
In supervised machine learning, we are given a collection of $n$ {\em training data} pairs $\left\{ (\bx_i, \by_i) \right\}_{i=1}^n$;
$\bx_i$ is termed the {\em data} and $\by_i$ the {\em label}.
Without loss of generality, we take $\bx_i\in\R^D,\by_i\in\R^C$ to be column vectors, but in practice they are often tensors.

We seek a {\em predictor} or {\em model} $f$ with two basic properties.
First, the predictor should {\em fit} the training data: $f(\bx_i)\approx \by_i$.
When the predictor fits (near) perfectly, we say that it has {\em interpolated} the data.
Second, the predictor should {\em generalize} to unseen data: $f(\bx')\approx \by'$, where $(\bx', \by')$ is {\em test data} that does not appear in the training set.
When we fit the training data but do not generalize, we say that we have {\em overfit}.

One solves the prediction problem by first designing a parameterized model $f_\Theta$ with parameters $\Theta$ and then {\em learning} or {\em training} by optimizing $\Theta$ to make $f_\Theta(\bx_i)$ as close as possible to $\by_i$ on average in terms of some distance or {\em loss} function ${\cal L}(\Theta)$ that measures the {\em training error}.

\noindent{\bf Deep networks.}
A {\em deep network} is a predictor or model constructed from the composition of $L$ intermediate mappings called {\em layers} \cite{goodfellow2016deep}
\begin{equation}
f_\Theta(\bx)= \left(f^{(L)}_{\theta^{(L)}} \circ \dots \circ f^{(1)}_{\theta^{(1)}}\right)\!(\bx).
\label{eq:layers}
\end{equation}
Here $\Theta$
is the collection of parameters from each layer, $\theta^{(\ell)}$, $\ell=1,\dots,L$.
We omit the parameters $\Theta$ or $\theta^{(\ell)}$ from our notation except when they are critical, since they are ever-present in the discussion below.

The $\ell$-th deep network layer $f^{(\ell)}$ takes as input the vector $\bz^{(\ell-1)}$ and outputs the vector $\bz^{(\ell)}$ by combining two simple operations:
\begin{equation}
\bz^{(\ell)} 
= f^{(\ell)}\left(\bz^{(\ell-1)}\right)
= \sigma \left(\bW^{(\ell)} \bz^{(\ell-1)} + \bb^{(\ell)} \right),
\label{eq:layer}
\end{equation}
where $\bz^{(0)}=\bx$ and $\bz^{(L)} = \widehat{\by} = f(\bx)$. 
First the layer applies an {\em affine transformation} to its input.
Second, in a standard abuse of notation, it applies a scalar nonlinear transformation --- called the {\em activation} function $\sigma$ --- to each entry in the result.
The entries of $\bz^{(\ell)}$ are called the layer-$\ell$ {\em neurons} or {\em units},
and the {\em width} of the layer is the dimensionality of $\bz^{(\ell)}$.
When layers of the form (\ref{eq:layer}) are used in (\ref{eq:layers}), deep learners refer to the network as a {\em multilayer perceptron} (MLP).

The parameters $\theta^{(\ell)}$ of the layer are the elements of the {\em weight matrix} $\bW^{(\ell)}$ and the {\em bias vector} $\bb^{(\ell)}$.
Special network structures have been developed to reduce the generally quadratic cost of multiplying by $\bW^{(\ell)}$. 
One notable class of networks constrains $\bW^{(\ell)}$ to be a circulant matrix, so that $\bW^{(\ell)}\bz^{(\ell)}$ corresponds to a convolution, giving rise to the term {\em ConvNet} for such models.
Even with this simplification, it is common these days to work with networks with many billions of parameters.

The most widespread activation function in modern deep
networks is the {\em rectified linear unit} (ReLU)
\begin{equation}
\sigma(u) = \max\{ u, 0 \} =: {\sf ReLU}(u).
\label{eq:ReLU}
\end{equation}
Throughout this paper, we focus on networks using this activation,
although the results hold for any continuous piecewise linear nonlinearity (e.g., absolute value, $\sigma(u)=|u|$).
Special activations are often employed at the last layer $f^{(L)}$, from the linear activation $\sigma(u)=u$ to the {\em softmax} that converts a vector to a probability histogram \cite{goodfellow2016deep}.
These activations do not affect our analysis below.
It is worth pointing out, but beyond the scope of this paper, that the results we review below extend to a much larger class of smooth activation functions (e.g., sigmoid gated linear units, Swish activation) by adopting a probabilistic viewpoint \cite{balestriero2018hard}.

\begin{figure}[h]
\centering
\includegraphics[width=1\linewidth]{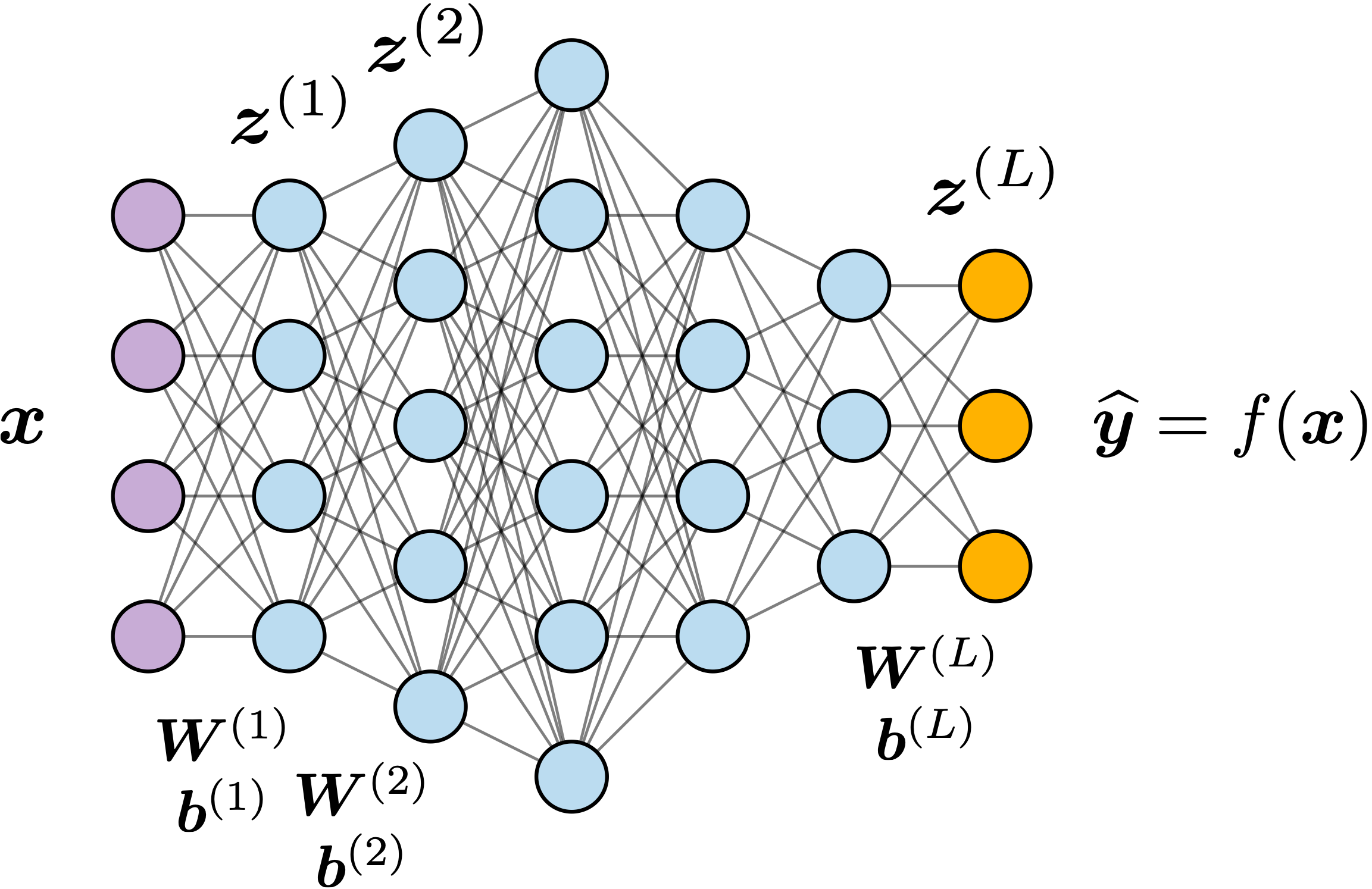}
\vspace*{-4mm}
    \caption{\small
    A 6-layer deep network.
    The purple, blue, and yellow nodes represent the input, neurons, and output, respectively, while the edges represent the affine transformation and activation effected by each layer.    
    The width of layer 2 is 5, for example.
    The links between the nodes represent the elements of the weight matrices $\bW^{(\ell)}$.
    The sum with the bias $\bb^{(\ell)}$ and subsequent  activation $\sigma(\cdot)$ are implicitly performed at each neuron.
    }
    \label{fig:deepNet}
\end{figure}

The term ``network'' is used in
deep learning because compositions of the form (\ref{eq:layers}) are often depicted as such; see Figure~\ref{fig:deepNet}.

\noindent{\bf Learning.}
To learn to fit the training data with a deep network, we tune the parameters $\bW^{(\ell)},\bb^{(\ell)}, \ell=1,\dots,L$ such that, on average, when training datum $\bx_i$ is input to the network, the output $\widehat{\by_i}=f(\bx_i)$ is close to $\by_i$ as measured by the loss function ${\cal L}$.
Two loss functions are ubiquitous in deep learning.
The first is the classical squared error based on the two-norm
\begin{equation}
{\cal L}(\Theta) :=
\frac{1}{n}
\sum_{i=1}^n \big\| \by_i - f_\Theta(\bx_i) \big\|_2^2
\label{eq:squared-loss}
\end{equation}
that is used in {\em regression} tasks, where the labels $\by_i$ are real-valued.
The other is the cross-entropy that is oft-used in classification tasks, where the labels are discrete.

Standard learning practice is to use some flavor of {\em gradient (steepest) descent} to iteratively reduce ${\cal L}$ by updating the parameters $\bW^{(\ell)},\bb^{(\ell)}$ by subtracting a small scalar multiple of the partial derivatives of ${\cal L}$ with respect to those parameters.

In practice, since the number of training data pairs $n$ can be enormous, one calculates the gradient of ${\cal L}$ for each iteration using only a subset of training data points and labels called a {\em minibatch}.

Note that even a nice loss function like (\ref{eq:squared-loss}) has a multitude of local minima due to the nonlinear activation $\sigma$ in each layer coupled with the composition of multiple such layers.
Consequently, numerous heuristics have been developed to help navigate to high-performing local minima.

In modern deep networks, the number of neurons is usually so gigantic that, by suitably optimizing the parameters, one can nearly interpolate the training data.
(We often drop the ``nearly'' below for brevity.)
What distinguishes the performance of one deep network architecture from another, then, is what it does away from the training points, i.e., how well it generalizes to unseen data.

\noindent{\bf Deep nets break out.} 
Despite neural networks existing in some form for over 80 years, their success was limited in practice until the AI boom of 2012. 
Sudden rapid progress was enabled by three converging factors: 
i) going deep with many layers (i.e., big $L$), ii) training on enormous data sets (i.e., big $n$), 
and iii) new computing architectures based on graphics processing units (GPUs). 

The spark that ignited the AI boom was the Imagenet Challenge 2012, where teams competed to best classify a set of input digital photographs into one of 1000 categories.
The Imagenet training data comprised about $n=1.3$ million, $D=$ 150,000-pixel color digital images human-labeled into $C=1000$ classes, such as `bird,' `bus,' `sofa.'
2012 was the first time a deep network won the Challenge; {\em AlexNet}, a ConvNet with 62 million parameters in five convolutional layers followed by three 
general layers achieved an accuracy of 60\%.
Subsequent competitions featured only deep networks, and, by the final  competition in 2017, they had reached 81\% accuracy, which is arguably better than most humans can achieve.

\noindent{\bf Black boxes.}
Deep networks with dozens of layers and millions or even billions of parameters are powerful for fitting and mimicking training data but also inscrutable. 
It is maddening that the mere composition of simple transformations (i.e., affine transforms and thresholding) so complicates analysis and defies detailed understanding.
Consequently, deep learning practitioners tend to treat them as black boxes and proceed empirically using an alchemical development process that focuses primarily on the inputs $\bx$ and outputs $f(\bx)$ of the network.
To truly understand deep networks we need to be able to see inside the black box as a deep network is learning and predicting.
In the sequel, we discuss one promising line of work in this vein that leverages the fact that deep networks are affine spline mappings.

\subsection*{Affine Splines}

\begin{figure}[h]
    \centering
\includegraphics[height=19mm]{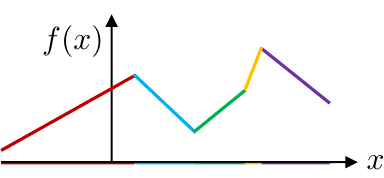} ~~
\includegraphics[height=19mm]{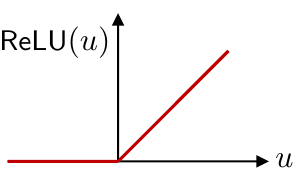}
    \caption{\small
    At left, a one-dimensional continuous piecewise linear function that we refer to as an affine spline. 
    At right, the ReLU activation function (\ref{eq:ReLU}) at the heart of many of today's deep networks.}
    \label{fig:1Dspline}
\end{figure}

As we now explain, deep networks are tractable multidimensional extensions of the familiar one-dimensional continuous piecewise linear functions depicted on the left in Figure~\ref{fig:1Dspline}.
When such continuous piecewise functions are fit to training data, we refer to them as {\em affine splines} for short.

Deep networks implement one particular extension of the affine spline concept to a multidimensional domain and range.
As we will see in the next section, a deep network generalizes the intervals of the independent variable over which a piecewise affine function is purely affine (recall Figure~\ref{fig:1Dspline}) to an irregular {\em tessellation} (tiling) of the network's $D$-dimensional input space into {\em convex polytopes}.
Let $\Omega$ denote the tessellation and $\omega\in\Omega$ an individual tile.
(The deep learning jargon for the polytope tiles is ``linear regions'' \cite{montufar2014number}.)

\begin{figure}[h]
    \centering
\includegraphics[width=0.8\linewidth]{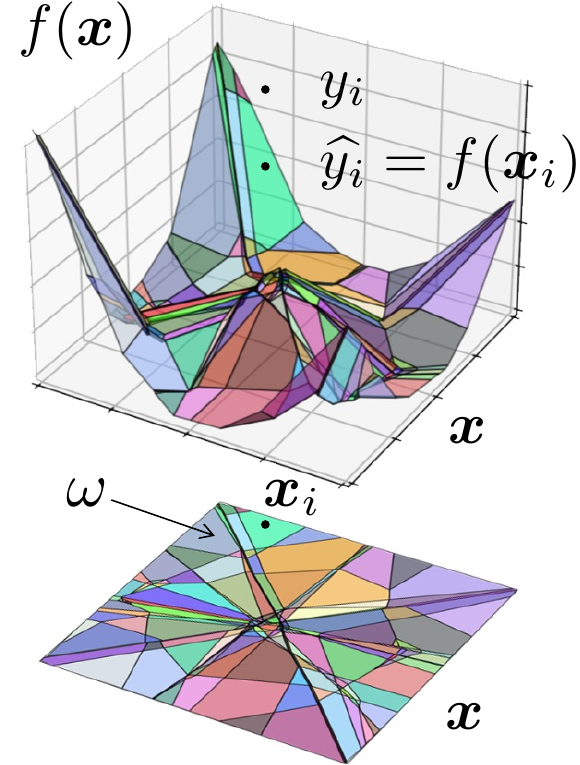}
    \caption{\small
    Input space tessellation $\Omega$ of the two-dimensional input space (below) and affine spline mapping $f(\bx)$ (above) for a toy deep network of depth $L=4$ and width 20.
    Also depicted is a training data pair $(\bx_i,y_i)$ and the prediction $\widehat{y_i}$.}
    \label{fig:CPA2D}
\end{figure}

Generalizing the straight lines defining the function on each interval in Figure~\ref{fig:1Dspline}, a deep network creates an affine transformation on each tile such that the overall collection is continuous.
Figure~\ref{fig:CPA2D} depicts an example for a toy deep network with a two-dimensional input space; here the tiles are polygons.
This all can be written as \cite{balestriero2021madmax}
\begin{equation}
f(\bx) = \sum_{\omega \in \Omega}(\bA_{\omega}\bx + \bc_{\omega}) \mathds{1}_{\{\bx \in \omega\}},
\label{eq:spline1}
\end{equation}
where the matrix $\bA_{\omega}$ and vector $\bc_{\omega}$ define the affine transformation from tile $\omega$ to the output.

Both the tessellation $\Omega$ and 
$\bA_{\omega},\bc_{\omega}$ from the affine transformations are functions of the deep network weights $\bW^{(\ell)}$ and biases $\bb^{(\ell)}$.
Geometrically, envision Figure~\ref{fig:CPA2D} with a cloud of $n$ training data points $(\bx_i,y_i)$;\footnote
{We remove the boldface from the labels in this example because they are scalars.} 
learning uses optimization to adjust the weights and biases to create a tessellation and affine transformations such that the affine spline predictions $\widehat{y_i}$ come as close as possible to the true labels $y_i$ as measured by the squared error loss (\ref{eq:squared-loss}), for example. 

One may wonder why we would set up this indirect deep network machinery just to implement an affine spline.
The reason is that direct spline representations are entirely impractical in machine learning settings, for two reasons.
First, we want the polytope tile boundaries to be unconstrained, and even for $D=1$, such ``free-knot'' splines are combinatorially complex to optimize.
Second, it is not clear how to extend the idea of a free-knot spline to even $D=2$, let alone to high dimensions.

\subsection*{Deep Network Tessellation}

As promised, let us now see how a deep network creates its input space tessellation \cite{balestriero2021madmax}.
Without loss of generality, we start with the first layer $f^{(1)}$ whose input is $\bx$ and output is $\bz^{(1)}$.
The $k$-th entry in $\bz^{(1)}$ (the value of the $k$-th neuron) is calculated simply as 
\begin{equation}
z^{(1)}_k= \sigma \left( 
\bw^{(1)}_k {\boldsymbol{\cdot}}\, \bx + b_k^{(1)} 
  \right),
  \label{eq:neuron}
\end{equation}
where the dot denotes the inner product, $\bw^{(1)}_k$ is the $k$-th row of the weight matrix $\bW^{(1)}$, and $\sigma$ is the ReLU activation function (\ref{eq:ReLU}).
The quantity inside the activation function is the equation of a $D-1$-dimensional hyperplane in the input space $\R^D$ that is perpendicular to $\bw^{(1)}_k$ and offset from the origin by $b_k^{(1)}/\| \bw^{(1)}_k \|_2$.
This hyperplane bisects the input space into two half-spaces; one where $z^{(1)}_k>0$ and one where $z^{(1)}_k=0$.

\begin{figure}[h]
    \centering
\includegraphics[width=0.8\linewidth]{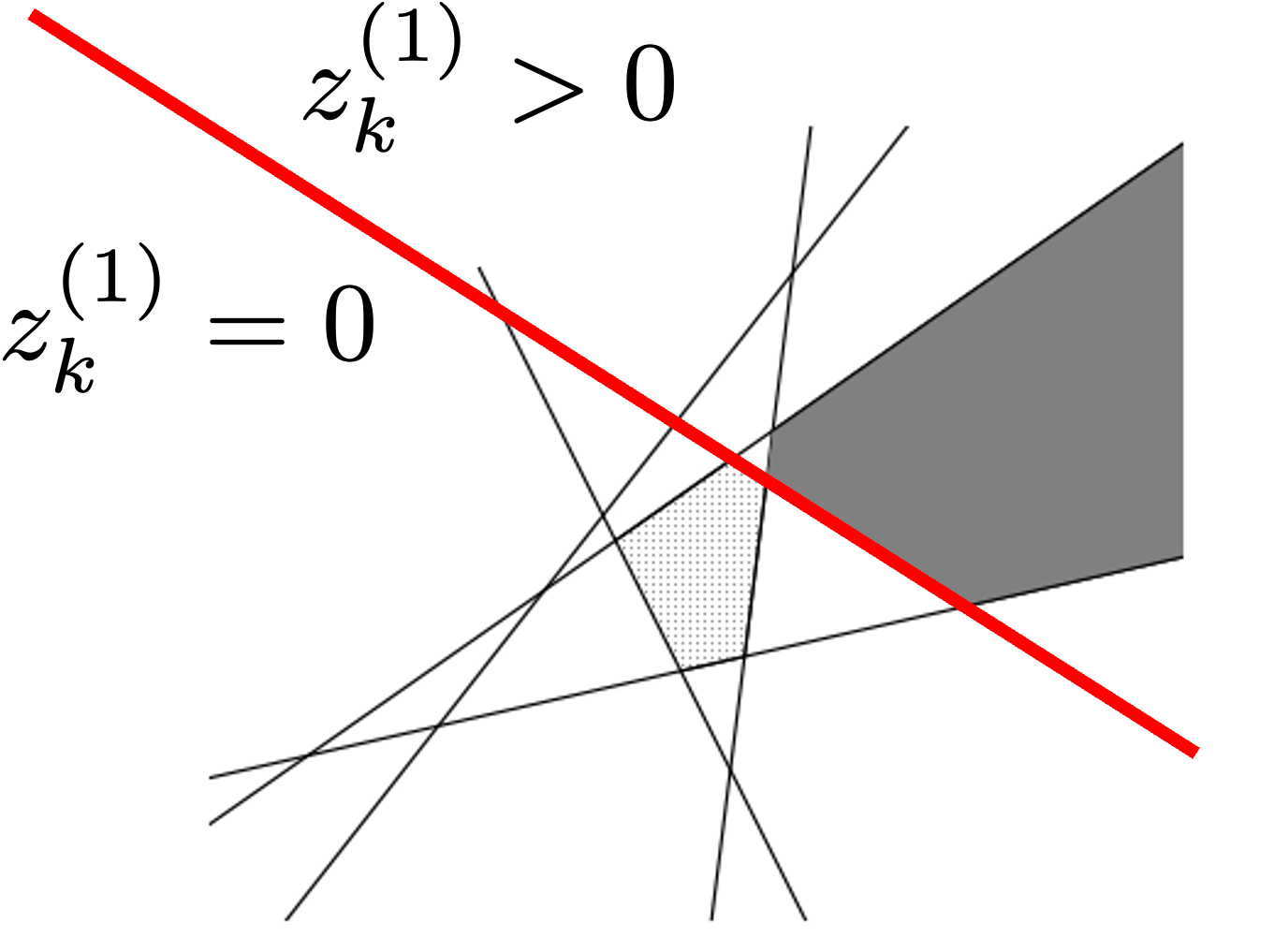}
    \caption{\small
    A deep network layer tessellates its input space into convex polytopal tiles via a hyperplane arrangement, with each hyperplane corresponding to one neuron at the output of the layer.
    In this two-dimensional example assuming ReLU activation, the red line indicates the one-dimensional hyperplane corresponding to the $k$-th neuron in the first layer.}
    \label{fig:ha}
\end{figure}

The collection of hyperplanes corresponding to each neuron in $\bz^{(1)}$ create a {\em hyperplane arrangement}.
It is precisely the intersections of the half-spaces of the hyperplane arrangement that tessellate the input space into convex polytope tiles (see Figure~\ref{fig:ha}).

The weights and biases of the first layer determine not only the tessellation of the input space but also an affine transformation on each tile to implement (\ref{eq:spline1}).
Explicit formulas for $\bA_\omega,\bc_\omega$ are available in \cite{romain}.
It should be clear that, since all of the transformations in (\ref{eq:neuron}) are continuous, so must be the affine spline (\ref{eq:spline1}) corresponding to the layer.

The tessellation corresponding to the composition of two or more layers follows an interesting {\em subdivision} process akin to a ``tessellation of tessellations'' \cite{romain}.
For example, the second layer creates a hyperplane arrangement in its input space, which happens to be the output space of layer one.
These hyperplanes can be pulled back through layer one to its input space by performing the same process as above but on a tile-by-tile basis relative to the layer-one tessellation and its associated affine transforms.
The effect on the layer-two hyperplanes is that they are {\em folded} each time they cross a hyperplane created by layer 1.
Careful inspection of the tessellation in Figure~\ref{fig:CPA2D} reveals many examples of such hyperplane folding.
Similarly, the hyperplanes created by layer three are folded every time they encounter a hyperplane in the input space from layers one or two.

Much can be said about this folding process, including a formula for the dihedral angle of a folded hyperplane as a function of the network's weights and biases \cite{romain}. 
Interestingly, the collection of dihedral angles of the folded hyperplanes from the final layer of a classification network determines the smoothness of the network's decision boundaries that partition the input space into regions corresponding to the $C$ classes.
Unfortunately, the formulae for the angles and affine transformations become unwieldy for more than two layers.
Finding simplifications for these attributes is an interesting open problem as are the connections to other subdivision processes like wavelets and fractals (more on this below).

The theory of hyperplane arrangements is rich and tells us that, generally speaking, the number of tiles grows 
rapidly with the number of neurons in each layer.
Hence, we can expect even modestly sized deep networks to have an enormous number of tiles in their input space, each with a corresponding affine transformation from the input to output space.
Importantly, though, the affine transformations are highly coupled because the overall mapping (\ref{eq:spline1}) must remain continuous.
This means that the class of functions that can be represented using a deep network is considerably smaller than if the mapping could be uncoupled and/or discontinuous.
Understanding what deep learning practitioners call the network's ``implicit bias'' remains an important open problem.

\subsection*{Visualizing the Tessellation}

The toy, low-dimensional examples in Figures~\ref{fig:CPA2D} and~\ref{fig:ha} are useful for building intuition, but how can we gain insight into the tessellation of a deep network with thousands or more of input and output dimensions? 
One way to proceed is to compute summary statistics about the tessellation, such as how the number of tiles scales as we increase the width or depth of a network (e.g., \cite{montufar2014number}); more on this below.
An alternative is to gain insight via direct visualization.

{\em SplineCam} is an exact method for computing and visualizing a deep network's spline tessellation over a specified low-dimensional region of the input space, typically a bounded two-dimensional planar slice \cite{splinecam}.
SplineCam uses an efficient graph data structure to encode the intersections of the hyperplanes from the various layers that pass through the slice and then uses a fast heuristic breadth-first search algorithm to identify tiles from the graph. 
All of the computations besides the search can be vectorized and computed on GPUs to enable the visualization of even industrial-scale deep networks.

\begin{figure}[h]
    \centering
\includegraphics[width=0.85\linewidth]{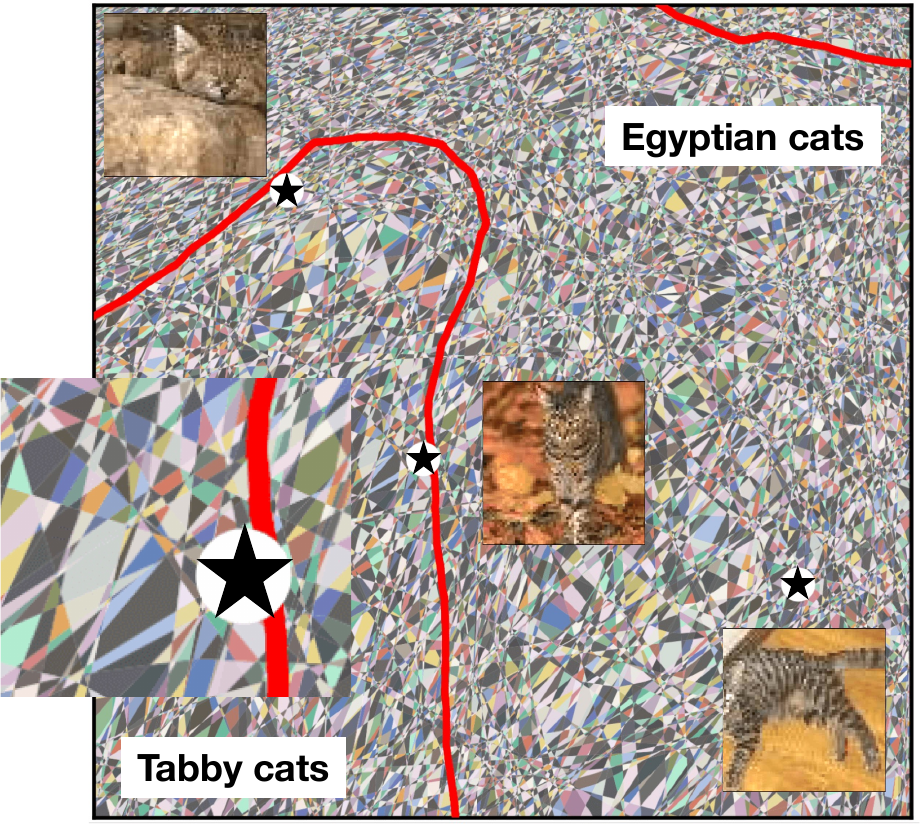}
    \caption{\small
    SplineCam visualization of a two-dimensional slice through the affine spline tessellation of the 4096-dimensional input space of a 5-layer ConvNet of average width 160 trained to classify $64 \times 64$ digital photos of cats. 
    The stars denote the three training images that define the plane and the red lines the decision boundaries between the two classes.
    (Adapted from \cite{splinecam}.)
}
    \label{fig:cats}
\end{figure}

Figure~\ref{fig:cats} depicts a SplineCam slice along the plane defined by three training images for a 5-layer ConvNet trained to classify between Egyptian and Tabby cat photos.
The first thing we notice is the extraordinarily large number of tiles in just this small region of the 4096-dimensional input space.
It can be shown that the decision boundary separating Egyptian and Tabby cats corresponds to a single hyperplane from the final layer that is folded extensively from being pulled back through the previous four layers \cite{romain}.
Photos falling in the lower left of the slice are classified as Tabbies, while photos falling in the lower right are classified as Egyptians. 
The density of tiles also varies across the input space.

An interesting avenue for future research involves the efficient extension of SplineCam to higher-dimensional slices both for visualization and the computation of summary statistics.

The main goal of this paper is to demonstrate the broad range of insights that can be garnered into the inner workings of a deep network through a focused study of the geometry of its input space tessellation.
To this end, we now tour five examples relating to deep network approximation, optimization, and data synthesis.
But we would be remiss if we did not point to the significant progress that has been made leveraging other important aspects of the spline view of deep learning, such as 
understanding how affine splines emerge naturally from the regularization typically used in deep network optimization \cite{unser2019representer} 
and 
what types of functions are learned by deep networks \cite{Parhi_2022}.


\subsection*{The Self-Similar Geometry of the \\ Tessellation}


It has been known since the late 1980s that even a two-layer neural network is a {\em universal approximator}, meaning that, as the number of neurons grows, one can approximate to arbitrary precision an arbitrary continuous function over a Borel measurable set  \cite{cybenko}. 
But, unfortunately, while two-layer networks are easily capable of interpolating a set of training data, in practice they do a poor job generalizing to data outside of the training set.
In contrast, deep networks with $L\gg 2$ layers have proved over the past 15 years that they are capable of both interpolating and generalizing well.

Several groups have investigated the connections between a network's depth and its tessellation's capacity to better approximate.
\cite{montufar2014number} was the first to quantify the advantage of depth by counting the number of tiles and showing that deep networks create more tiles (and hence are more expressive) than shallow networks.

Furthermore, deeper networks realize more folded hyperplanes, which can impart more nonlinearity in the spline mapping than a shallower network.
For instance, the hyperplane corresponding to a neuron from the first layer can only linearly divide the input space into two half-spaces, while a later-layer neuron can carve up the input space into myriad, even disconnected regions \cite{romain}.

Additional work has worked to link the self-similar nature of the tessellation to good approximation.
Using self-similarity, one can construct new function spaces for which deeper networks provide better approximation rates (see \cite{devore2021neural,daubechies2022nonlinear} and the references therein).
The benefits of depth stem from the fact that the model is able to replicate a part of the function it is trying to approximate in many different places in the input space and with different scalings or orientations.
Extending these results, which currently hold only for one-dimensional input and output spaces, to multidimensional signals is an interesting open research avenue.
The subdivision results from \cite{romain} could prove useful here.

\subsection*{Geometry of the Loss Function}

Frankly, it seems an apparent miracle that deep network learning even works.
Because of the composition of nonlinear layers and the myriad local minima of the loss function, deep network optimization remains an active area of empirical research.
Here we look at one analytical angle that exploits the affine spline nature of deep networks.

Over the past decade, a menagerie of different deep network architectures has emerged that innovate in different ways on the basic architecture (\ref{eq:layers}), (\ref{eq:layer}).
A natural question for the practitioner is: Which architecture should be preferred for a given task?
Approximation capability does not offer a point of differentiation, because, as we just discussed, as their size (number of parameters) grows, most deep networks attain a universal approximation capability. 

\begin{figure}[h]
    \centering  
\includegraphics[width=1\linewidth]{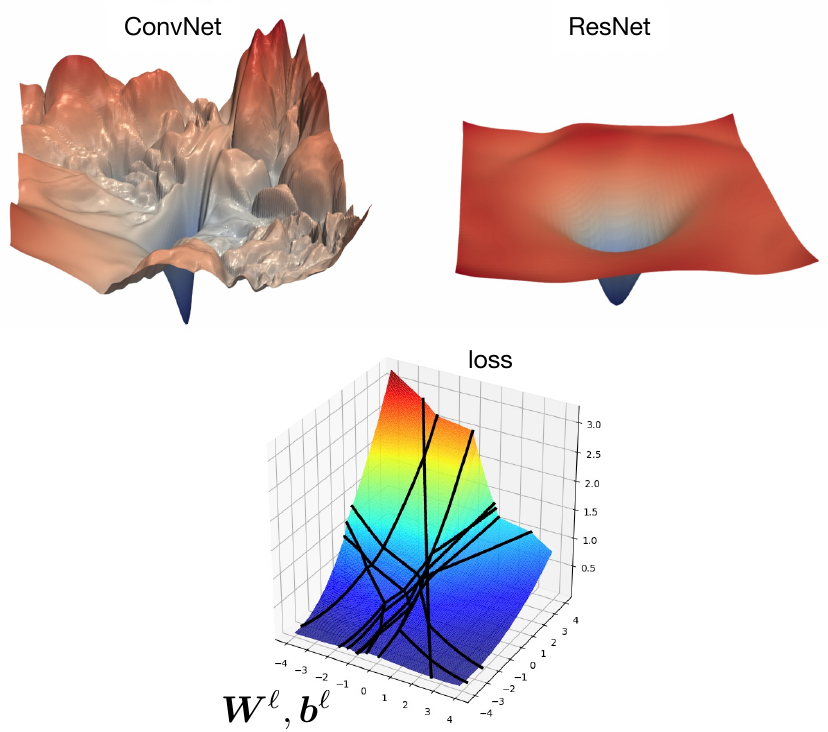}
\vspace*{-5mm}
    \caption{\small
    Loss landscape ${\cal L}(\Theta)$ of a ConvNet and ResNet (from \cite{li2018visualizing}). Piecewise quadratic loss function.}
    \label{fig:loss_surface}
\end{figure}

Practitioners know that deep networks with {\em skip connections}
\begin{equation}
\bz^{(\ell)} = \sigma\left(\bW^{(\ell)} \bz^{(\ell-1)} + \bb^{(\ell)} \right) + \bz^{(\ell-1)}
\label{eq:skip}
\end{equation}
such as so-called {\em ResNets}, are much preferred over ConvNets, because empirically their gradient descent learning converges faster and more stably to a better minimum.
In other words, it is not {\em what} a deep network can approximate that matters, but rather {\em how it learns} to approximate. 
Empirical studies have indicated that this is because the so-called {\em loss landscape} of the loss function ${\cal L}(\Theta)$ navigated by gradient descent as it optimizes the deep network parameters is much smoother for ResNets as compared to ConvNets (see Figure~\ref{fig:loss_surface}).
However, to date there has been no analytical work in this direction.

Using the affine spline viewpoint, it is possible to analytically characterize the local properties of the deep network loss landscape and  quantitatively compare different deep network architectures.
The key is that, for the deep networks under our consideration trained by minimizing the squared error (\ref{eq:squared-loss}), the loss landscape ${\cal L}$ as a function of the deep network parameters $\bW^{(\ell)},\bb^{(\ell)}$ is a {\em continuous piecewise quadratic function}  
\cite{rolf,ankit2020} 
that is amenable to analysis (see Figure~\ref{fig:loss_surface}).

The optimization of quadratic loss surfaces is well-understood.
In particular, the eccentricity of a quadratic loss landscape is governed by the {\em singular values} of the Hessian matrix containing the second-order quadratic terms.
Less eccentric (more bowl shaped) losses are easier for gradient descent to quickly navigate to the bottom. 
Similarly, the local eccentricity of a continuous piecewise quadratic loss function and the width of each local minimum basin are governed by the singular values of a ``local Hessian matrix'' that is a function of not only the deep network parameters but also the deep network architecture.
This enables us to quantitatively compare different deep network architectures in terms of their singular values.

In particular, we can make a fair, quantitative comparison between the loss landscapes of the ConvNet and ResNet architectures by comparing their singular values. 
The key finding is that the condition number of a ResNet (the ratio of the largest to smallest singular value) is bounded, while that of the ConvNet is not \cite{rolf}.
This means that the local loss landscape of a ResNet with skip connections is provably better conditioned than that of a ConvNet and thus less erratic, less eccentric, and with local minima that are more accommodating to gradient descent optimization.

Beyond analysis, one interesting future research avenue in this direction is converting this analytical understanding into new optimization algorithms that are more efficient than today's gradient descent approaches.

\subsection*{The Geometry of Initialization }

As we just discussed, even for the prosaic squared error loss function (\ref{eq:squared-loss}), the loss landscape as a function of the parameters is highly nonconvex with myriad local minima.
Since gradient descent basically descends to the bottom of the first basin it can find, where it starts (the initialization) really matters.
Over the years, many techniques have been developed to improve the initialization and/or help gradient descent find better minima; here we look at one of them that is particularly geometric in nature.

With {\em batch normalization}, 
we modify the definition of the neural computation from (\ref{eq:neuron}) to 
\begin{equation}
    z^{(1)}_k=
    \sigma\left(
    \frac{\bw^{(1)}_k {\boldsymbol{\cdot}}\, \bx 
    -\mu^{(1)}_k} {\nu^{(1)}_k}
    \right),
    \label{eq:batch normalization}
\end{equation}
where $\mu^{(1)}_k$ and $\nu^{(1)}_k$ are not learned by gradient descent but instead are directly computed as the mean and standard deviation of $\bw^{(1)}_k \boldsymbol{\cdot}\, \bx_i$ over the training data inputs involved in each gradient step in the optimization.
Importantly, this includes the very first step, and so batch normalization directly impacts the initialization from which we start iterating on the loss landscape.\footnote{As implemented in practice, batch normalization has two additional parameters that are learned as part of the gradient descent; however \cite{BN-arxiv} shows that these parameters have no effect on the optimization initialization and only a limited effect during learning as compared to $\mu$ and $\nu$.}

Astute readers might see a connection to the standard statistical data preprocessing step of data normalization and centering; the main difference is that this processing is performed before each and every gradient learning step.
Batch normalization often greatly aids the optimization of a wide variety of deep networks, helping it to find a better (lower) minimum quicker.
But the reasons for its efficacy are poorly understood.

We can make progress on understanding batch normalization by again leaning on the affine spline viewpoint.
Let's focus on the effect of batch normalization at initialization just before gradient learning begins; the effect is pronounced, and it is then easy to extrapolate regarding what happens at subsequent gradient steps.
Prior to learning, a deep network's weights are initialized with random values. 
This means that the initial hyperplane arrangement is also random.

The key finding of \cite{BN-arxiv} is that batch normalization adapts the geometry of a deep network's spline tessellation to focus the network's attention on the training data $\left\{ \bx_i \right\}_{i=1}^n$.
It does this by adjusting the angles and offsets of the hyperplanes that form the boundaries of the polytopal tiles to increase their density in regions of the input space inhabited by the training data, thereby enabling finer approximation there.

\begin{figure}[h]
    \centering  
    \begin{minipage}{0.49\linewidth}
    \centering
\includegraphics[width=\linewidth]{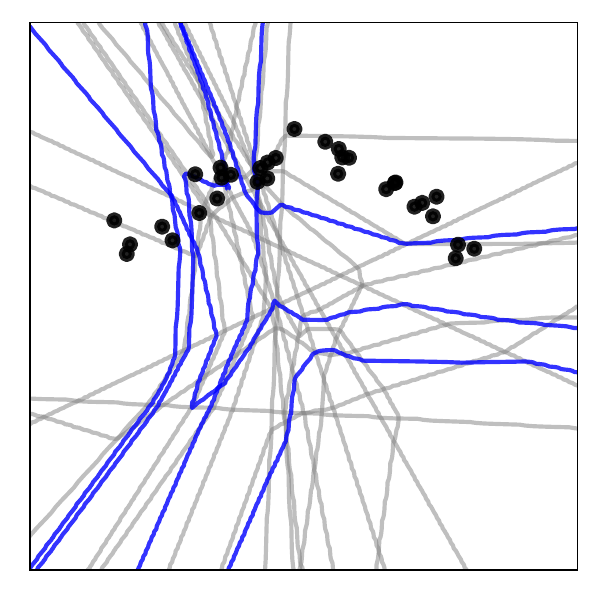}\\[0em]
{\small\sf without batch norm}
\end{minipage}
\begin{minipage}{0.49\linewidth}
    \centering
\includegraphics[width=\linewidth]{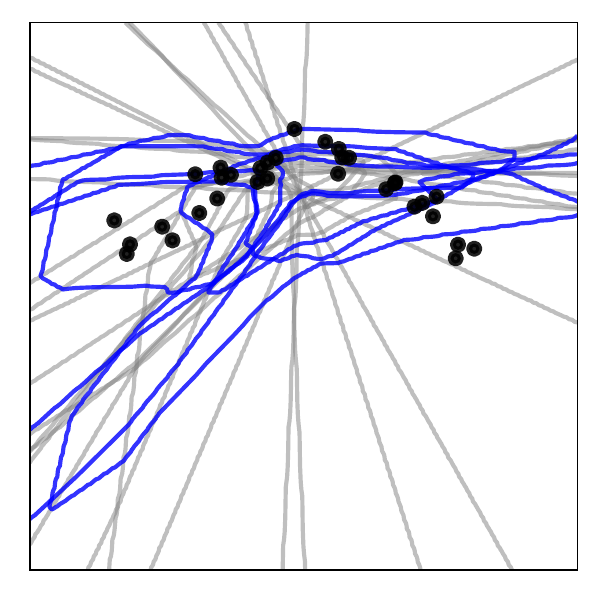} \\[0em]
{\small\sf with batch norm}
\end{minipage}
    \caption{\small
    Visualization of a set of two-dimensional data points $\bx_i$ (black dots) and the input-space spline tessellation of a 4-layer toy deep network with 
    random weights $\bW^{(\ell)}$.
    The grey lines correspond to (folded) hyperplanes from the first three layers.
    The blue lines correspond to folded hyperplanes from the fourth layer. 
    (Adapted from \cite{BN-arxiv}.)
}
    \label{fig:BN1}
\end{figure}

More precisely, batch normalization directly adapts each layer's input space tessellation to minimize the {\em total least squares distance} between the tile boundaries and the training data.
The resulting data-adaptive initialization aligns the spline tessellation with the data not just at initialization but before every gradient step to give the learning algorithm a much better chance of finding a quality set of weights and biases.
See Figure~\ref{fig:BN1} for a visualization.

\begin{figure}[h]
\centering
\setlength{\tabcolsep}{0pt}
\begin{tabular}{rcccc}
& {\small\sf data} & {\small $\ell=3$} & {\small $\ell=7$} & {\small $\ell=11$} \\
\rotatebox{90}{\small\sf ~~~zero bias}
\hspace*{0.5pt}
&
\includegraphics[width=0.23\linewidth]{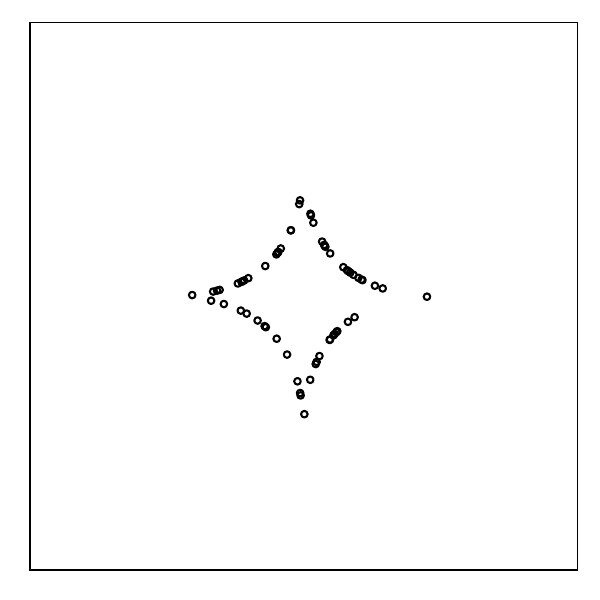}
& 
\includegraphics[width=0.23\linewidth]{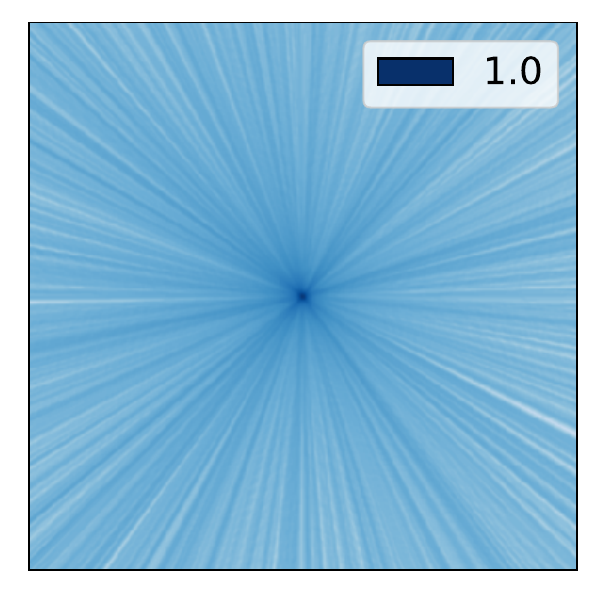}
& 
\includegraphics[width=0.23\linewidth]{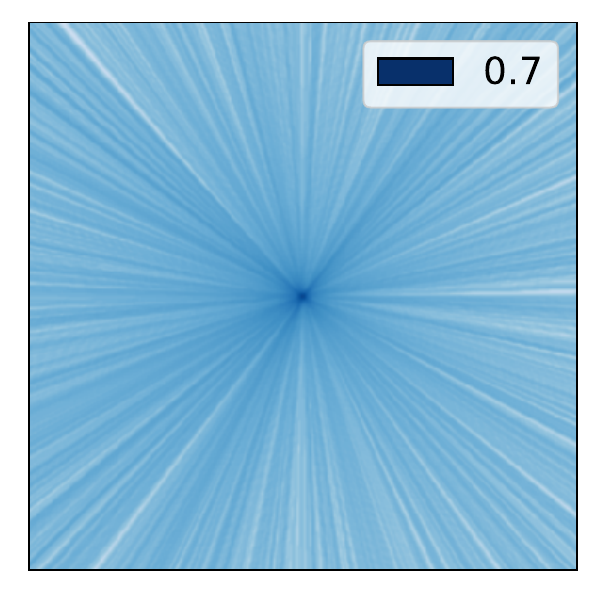}
& 
\includegraphics[width=0.23\linewidth]{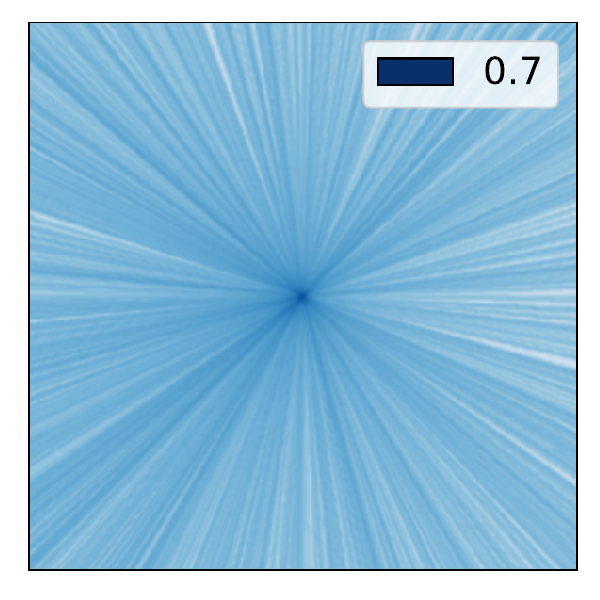}
\\[-3.5pt]
\rotatebox{90}{\small\sf ~random bias}
\hspace*{0.5pt}
&
\includegraphics[width=0.23\linewidth]{Figs/x_5.pdf}
& 
\includegraphics[width=0.23\linewidth]{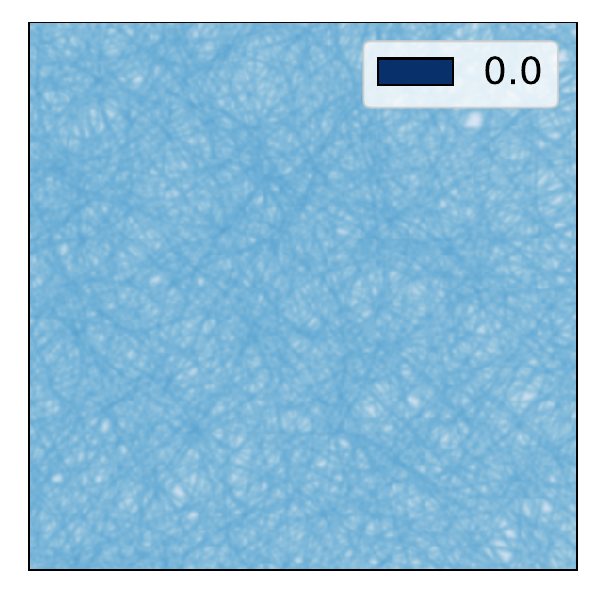}
& 
\includegraphics[width=0.23\linewidth]{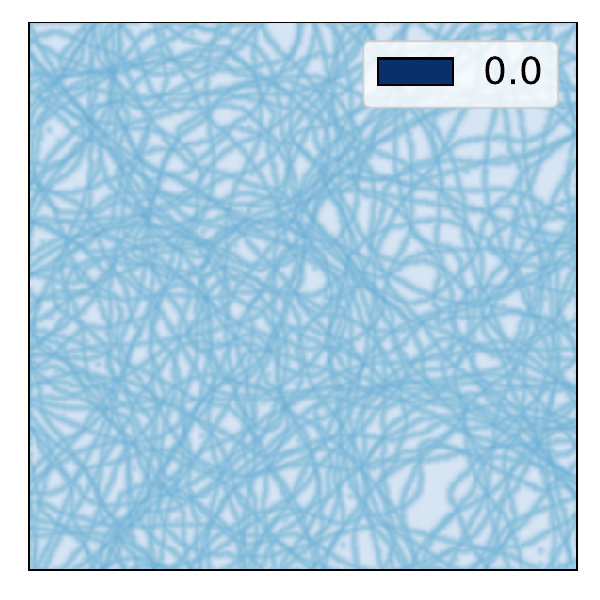}
& 
\includegraphics[width=0.23\linewidth]{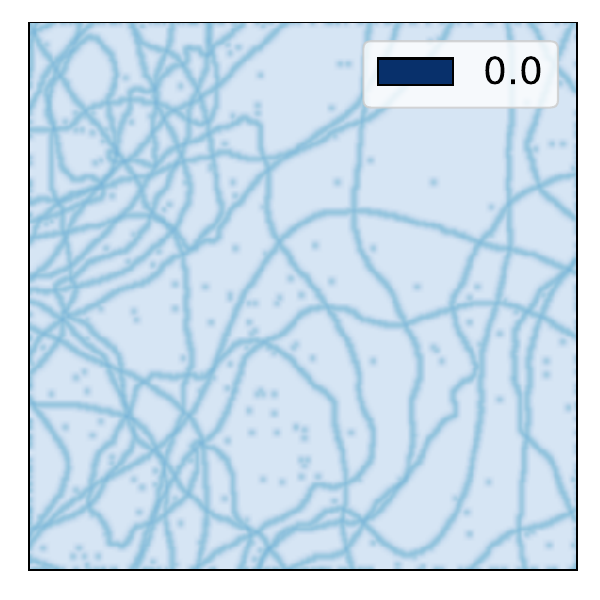}
\\[-3.5pt]
\rotatebox{90}{\small\sf ~batch norm} 
\hspace*{0.5pt}
&
\includegraphics[width=0.23\linewidth]{Figs/x_5.pdf}
& 
\includegraphics[width=0.23\linewidth]{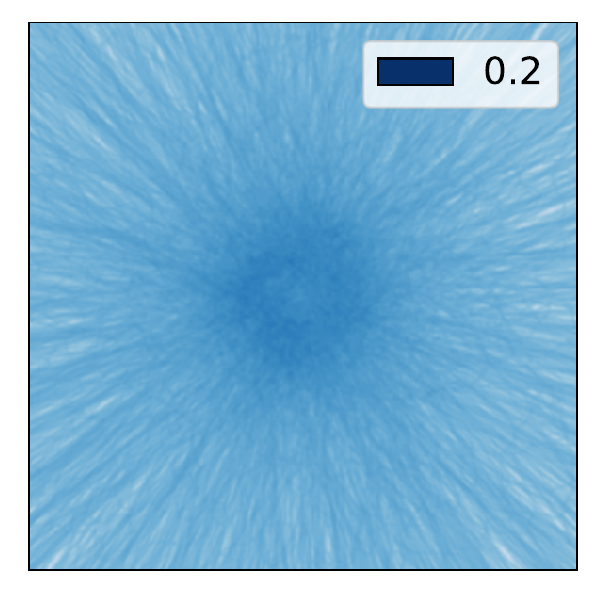}
& 
\includegraphics[width=0.23\linewidth]{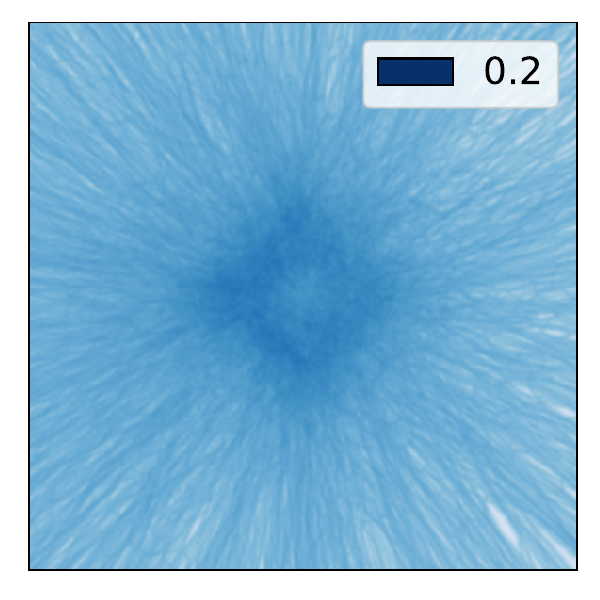}
& 
\includegraphics[width=0.23\linewidth]{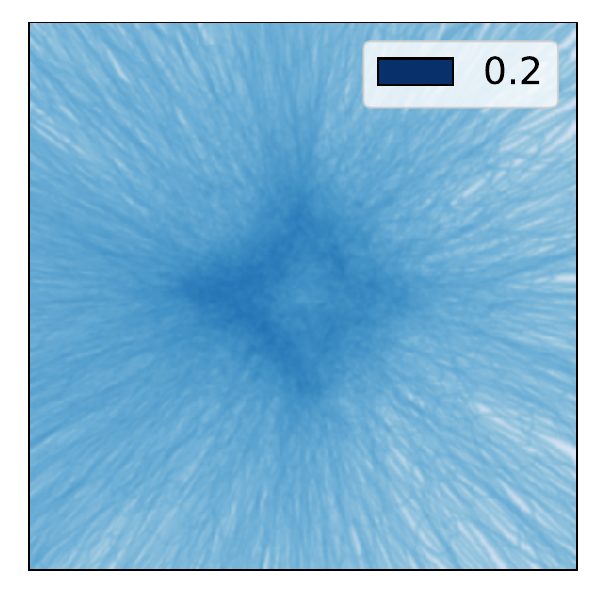}
\end{tabular}
    \caption{\small
Densities of the hyperplanes created by layers 3, 7, and 11 in the two-dimensional input space of an 11-layer deep network of width 1024.
The training data consists of 50 samples from a star-shaped distribution.
(Adapted from \cite{BN-arxiv}.)
  }
\label{fig:BN2}
\end{figure}

Figure~\ref{fig:BN2} provides clear evidence of batch normalization's adaptive prowess. 
We initialize an 11-layer deep network with a two-dimensional input space three different ways to train on data with a star-shaped distribution.
We plot the density of the hyperplanes (basically, the number of hyperplanes passing through local regions of the input space) created by layers 3, 7, and 11 for three different layer configurations:
i) the standard layer (\ref{eq:neuron}) with bias $\bb^{(\ell)}={\bf 0}$; 
ii) the standard layer (\ref{eq:neuron}) with random bias $\bb^{(\ell)}$;
iii) the batch normalization layer (\ref{eq:batch normalization}).
In all three cases, the weights $\bW^{(\ell)}$ were initialized randomly.
We can make several observations.
First, constraining the bias to be zero forces the network into a central hyperplane arrangement tessellation that is incapable of aligning with the data.
Second, randomizing both the weights and biases splays the tiles over the entire input space, including many places where the training data is not.
Third, batch normalization focuses the hyperplanes from all three of the layers onto the regions where the training data lives.

One interesting avenue for future research in this direction is developing new normalization schemes that replace the total least squares optimization to enforce a specific kind of adaptivity of the tessellation to the data and task at hand.


\subsection*{The Dynamic Geometry of Learning}


Current deep learning practice treats a deep network as a black box and optimizes its internal parameters (weights and biases) to minimize some end-to-end training error like the squared loss (\ref{eq:squared-loss}).
While this approach has proved mightily successful empirically, it provides no insight into how learning is going on inside the network nor how to improve it.
Clearly, as we adjust the parameters to decrease the loss function using gradient descent, the tessellation will change dynamically.
Can we use this insight to glean something new about what goes on inside a deep network during learning?

\begin{figure*}
\centering
\setlength{\tabcolsep}{0pt}
\begin{tabular}{ccc}
\includegraphics[width=0.33\linewidth]{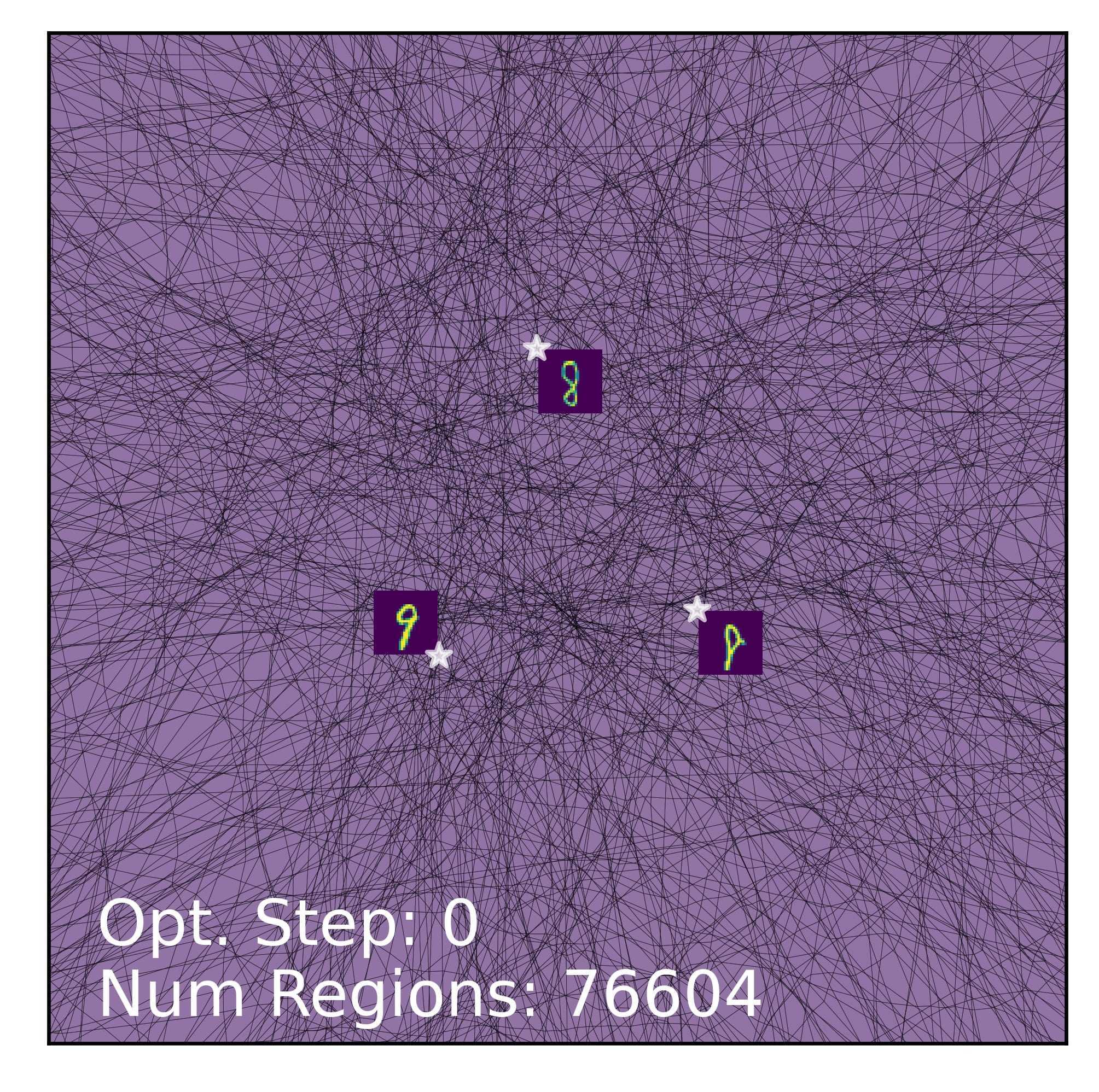}
& 
\includegraphics[width=0.33\linewidth]{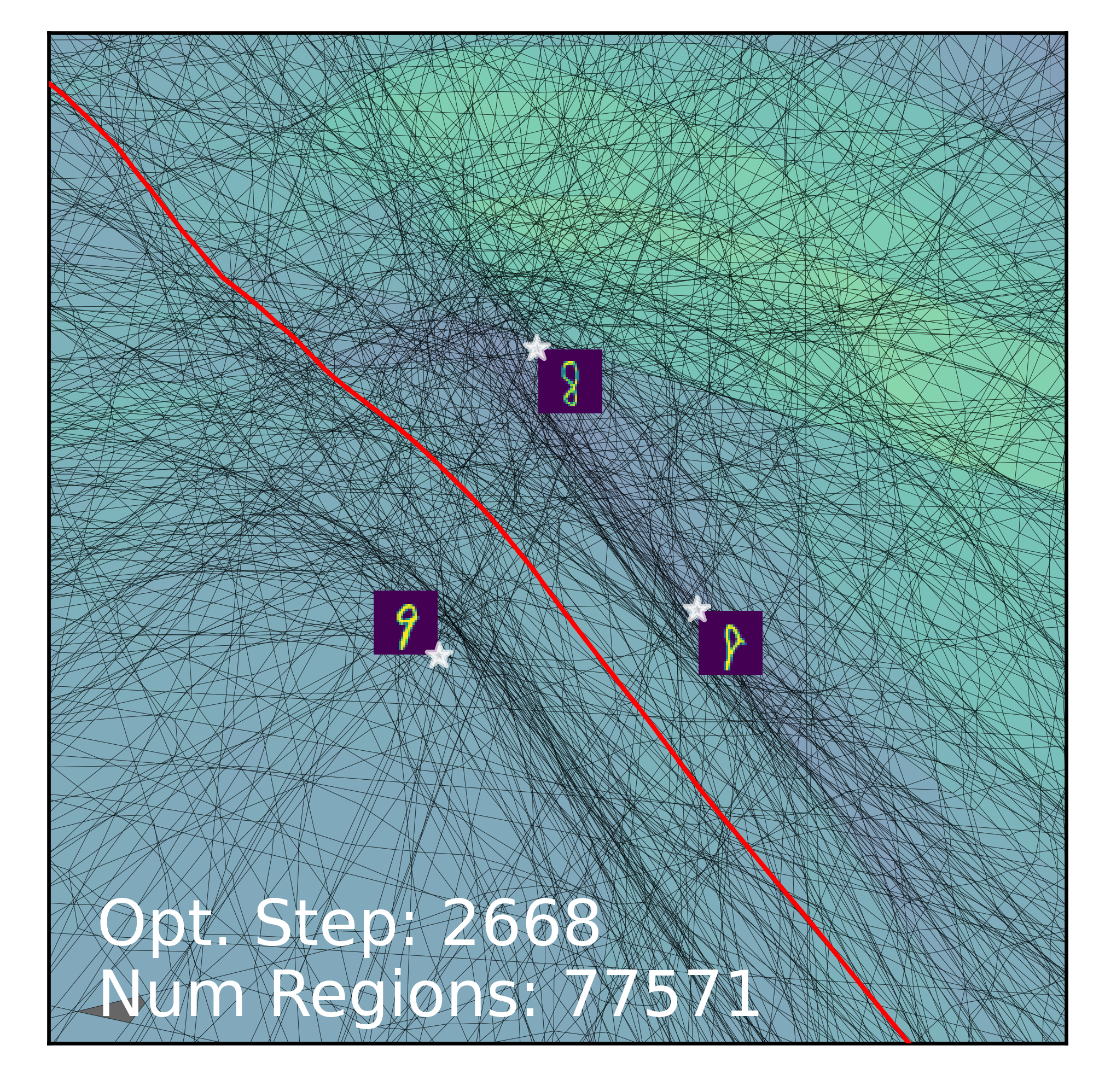}
& 
\includegraphics[width=0.33\linewidth]{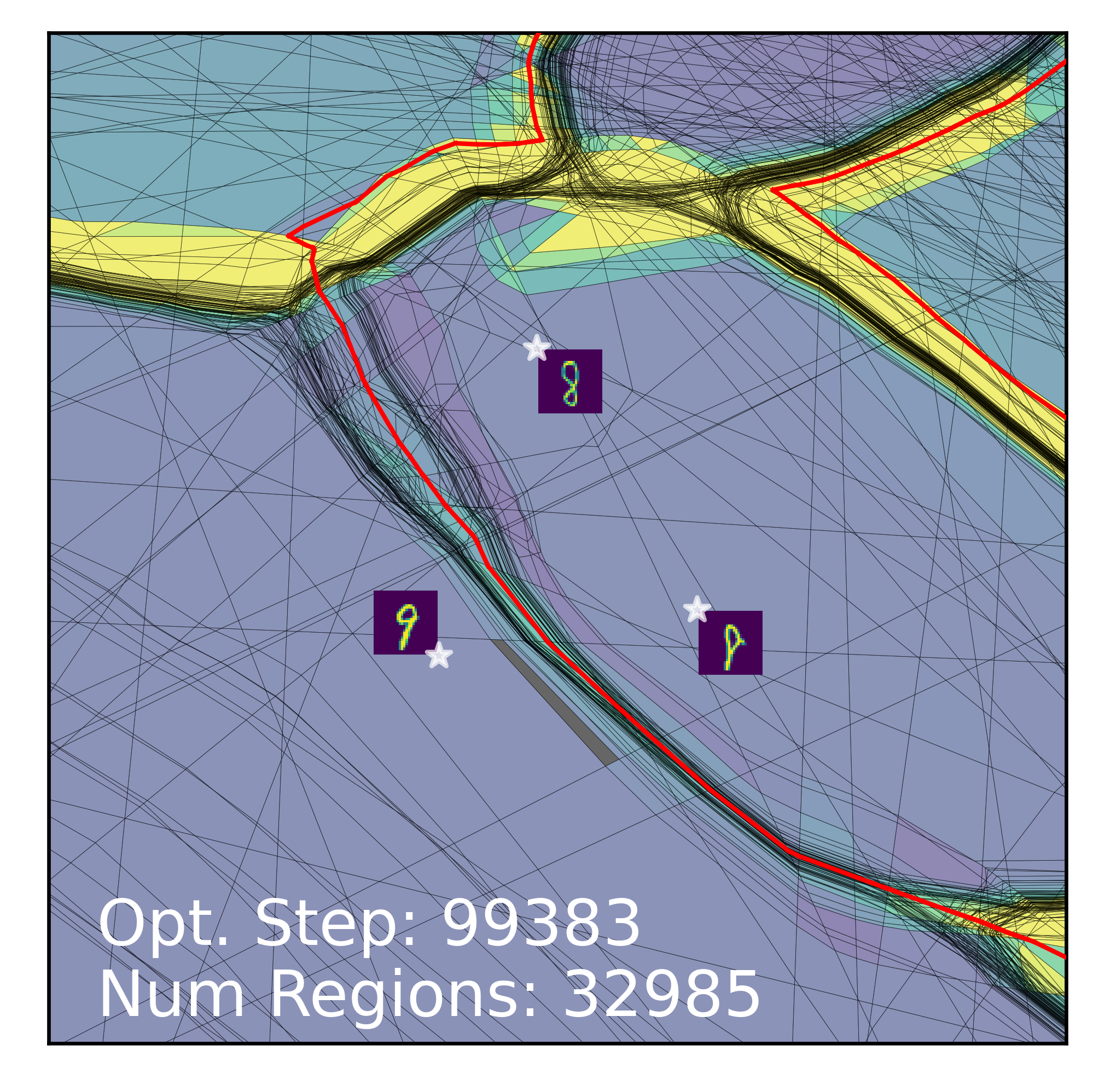} 
\\[-1mm]
{\scriptsize ${\sl LC}=4.91$} & {\scriptsize ${\sl LC}=2.96$} & {\scriptsize ${\sl LC}=0.142$}
\\
{\small\sf initialization} & {\small\sf interpolation} & {\small\sf grokking}
\end{tabular}
    \vspace*{-3mm}
    \caption{\small
    SplineCam visualization of a slice of the input space defined by three training MNIST digits being classified by a 4-layer MLP of width 200.
    The false color map (vivirdis) encodes the 2-norm of the $\bA_\omega$ matrix defined on each tile according to purple (low), green (medium), yellow (high). 
    The decision boundary is depicted in red. 
    (Adapted from \cite{alwaysgrok}.)
    }
    \label{fig:grok}
\end{figure*}

Consider a deep network learning to classify photos of the handwritten digits 0--9.
Figure~\ref{fig:grok} deploys SplineCam to visualize a portion of a 2D slice of the input space of the network defined by three data points in the MNIST handwritten digit training dataset \cite{alwaysgrok}.
At left, we see that the tessellation at {\em initialization} (before we start learning) is in disarray due to the random weights and biases and nonuse of batch normalization (more on this later).
The tessellation is random, and the training error is large.

In the middle, we see the tessellation after convergence to near-zero training error ({\em interpolation}), when most of the digits are on the correct side of their respective decision boundaries.
Not shown by the figure is the fact that the network also generalizes well to unseen test digits at this juncture.
High tile density suggests that even a continuous piecewise affine function can be quite rugged around these points \cite{sebastianParis}.
Indeed, the false coloring indicates that the 2-norms of the $\bA_\omega$ matrices has increased around the training images, meaning that their ``slopes'' have increased.
As a consequence, the overall spline mapping $f(\bx)$ is now likely more rugged and more sensitive to changes in the input $\bx$ as measured by a local (per-tile) Lipschitz constant.
In summary, at (near) interpolation, the gradient learning iterations have in some sense accomplished their task (near zero training error) but with elevated sensitivity of $f(\bx)$ to changes in $\bx$ around the training data points as compared to the random initialization.

Interpolation is the point where most deep learning practitioners would stop training and fix the network's weights and biases for use in their target application.
But let's see what happens if we continue training about 37 times longer.
At right in Figure~\ref{fig:grok}, we see that, while the training error does not improve after continued training (it is still near zero, meaning correct classification of nearly all the training data), the tessellation has metamorphosed.
There are now only half as many tiles in this region, and they have nearly all migrated to define the decision boundary, where presumably they are being used to create sharp decisions.
Around the training data, we have an extremely low density of tiles with low 2-norm of their $\bA_\omega$ matrices and thus presumably a much smoother mapping $f(\bx)$.
Hence, the sensitivity of $f(\bx)$ around the training data as measured by a local Lipshitz constant will be much lower than just after interpolation.

This situation is an example of {\em grokking}, in which some desired property of a deep network unexpectedly occurs well after the training error converges to near zero \cite{power2022grokking}. 
In this case, the desired property is the robustness of the network to perturbations in the training data.
A dirty secret of today's deep networks is that $f(\bx)$ can be quite unstable to small changes in $\bx$ (which seems expected given their high degree of nonlinearity).
This instability makes deep networks less robust and more prone to so-called {\em adversarial attacks} such as causing a `barn' image to be classified as a `pig' by adding a nearly undetectable but carefully designed attack signal to an image of a barn.
Continuing learning to achieve grokking and {\em delayed robustness} is a new approach to mitigating such attacks in particular and making deep learning more stable and predictable in general.


Can we translate the visualization of Figure~\ref{fig:grok} into a metric that can be put into practice to compare or improve deep networks?
This is largely an open research question, but here are some first steps \cite{alwaysgrok}. 
Define the {\em local complexity} (LC) as the number of tiles in a neighborhood $V$ around a point $\bx$ in the input space. 
While exact computation of the LC is combinatorially complex, an upper bound can be obtained in terms of the number of hyperplanes that intersect $V$ according to Zaslavsky’s Theorem, with the assumption that $V$ is small enough that the hyperplanes are not folded inside $V$. 
Therefore, we can use the number of hyperplanes intersecting $V$ as a proxy for the number of tiles in $V$.

For the experiment reported in Figure~\ref{fig:grok}, we computed the LC in the neighborhood of each data point in the entire training dataset and then averaged those values.
From the above discussion, high LC around a point $\bx$ in the input space implies a multitude of small tiles in that region and a potentially unsmooth and unstable mapping $f(\bx)$ around $\bx$.
The values reported in Figure~\ref{fig:grok} confirm that the LC does indeed capture the intuition that we garnered visually.

Open research questions regarding the dynamics of deep network learning abound.
At a high level, it is clear from Figure~\ref{fig:grok} that the classification function ultimately being learned has its curvature concentrated at the decision boundary.
Approximation theory would agree that a free-form spline should indeed concentrate its tiles around the decision boundary to minimize the approximation error. 
However, it is not clear why the migration occurs so late in the training process.

Another interesting research direction involves incorporating the average LC around the training data points in the optimization cost function (e.g., (\ref{eq:squared-loss})) in order to encourage the network to converge to a stable mapping $f(\bx)$ sooner rather than much later.


Yet another interesting research direction is the interplay between grokking and batch normalization, which we discussed above. 
Batch normalization provably concentrates the tessellation near the training data points, but (at least for classification problems) to grok we need the tiles to move away from those points towards the decision boundaries.
Hence, it is clear that batch normalization and grokking compete with each other. 
How to get the best of both worlds at both ends of the gradient learning timeline is an open question.


\subsection*{The Geometry of Generative Models}

A {\em generative model} aims to learn the underlying patterns in the training data in order to generate new, similar data. 
The current crop of deep generative models includes transformer networks that power large language models for text synthesis and chatbots and diffusion models for image synthesis.
Here we investigate the geometry of models that until recently were state-of-the-art, such as Generative Adversarial Networks (GANs) and Variational Autoencoders (VAEs) that are often based on ReLU and other piecewise linear activation functions.

Deep generative models typically map from a low-dimensional Euclidean input space (called the parameter space) to a {\em manifold} ${\cal M}$ of roughly the same dimension in a high-dimensional output space.
Each point $\bx$ in the parameter space synthesizes a corresponding output point $\widehat{\by}= f(\bx)$ on the manifold (e.g., a picture of a bedroom).
Training on a large number of data points $\by_i$ learns an approximation to the mapping $f$ from the parameter space to the manifold. 
It is beyond the scope of this review, but learning the parameters of a deep generative model is usually more involved than simple gradient descent \cite{goodfellow2016deep}.
It can be useful for both training and synthesis to view the points $\bx$ from the parameter space as governed by some probability distribution, e.g., uniform over a bounded region of the input space.

\begin{figure}[h]
    \centering  
\includegraphics[width=1\linewidth]{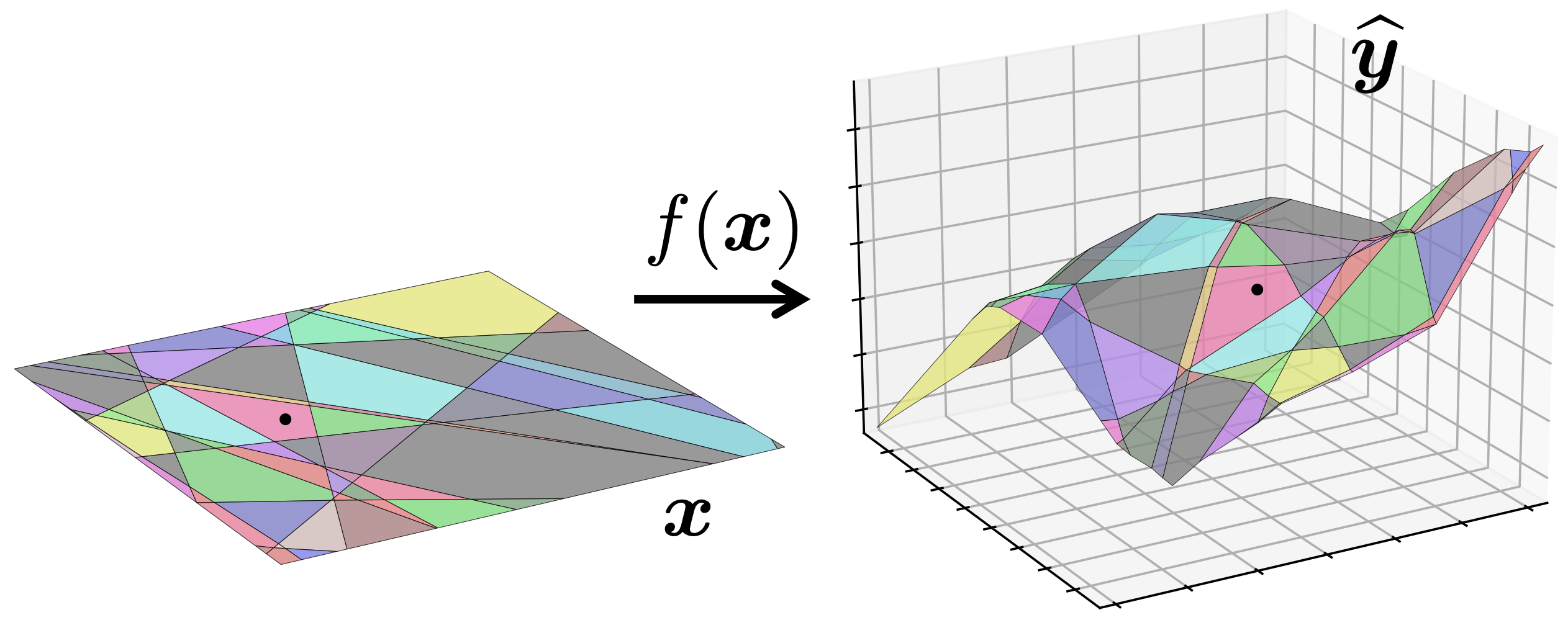}
\vspace*{-5mm}
    \caption{\small
    A ReLU-based deep generative network manifold ${\cal M}$ is continuous and piecewise affine.
    Each affine spline tile $\omega$ in the input space is mapped by an affine transformation to a corresponding tile $M(\omega)$ on the manifold.}
    \label{fig:manifold}
\end{figure}

For a deep generative model based on a ReLU or similar activation function, the manifold ${\cal M}$ is a {\em continuous piecewise affine manifold};\footnote{We allow ${\cal M}$ to intersect itself transversally in this setting.} see Figure~\ref{fig:manifold}.
Points on the manifold $f(\bx)$ are given by (\ref{eq:spline1}) as the parameter vector $\bx$ sweeps through the input parameter space.

A major issue with deep generative models is that, if the training data is not carefully sourced and currated, then they can produce biased outputs.
A deep generative model like a GAN or VAE is trained to approximate both the structure of the true data manifold from which the training data was sampled and the data distribution on that manifold. 
However, all too often in practice, training data are obtained based on preferences, costs, or convenience factors that produce artifacts in the training data distribution on the manifold.
Indeed, it is common in practice for there to be more training data points in one part of the manifold than another.
For example, a large fraction of the faces in the CelebA dataset are smiling, and a large fraction of those in the FFHQ dataset are female with dark hair.
When one samples uniformly from a model trained using such biased data, the biases 
are reproduced when sampling from the trained model, which has far-reaching implications for algorithmic fairness and beyond.

We can both understand and ameliorate sampling biases in deep generative models by again leveraging their affine spline nature.
The key insight for the bias issue is that the tessellation of the input parameter space is carried over onto the manifold.
That is, each convex tile $\omega$ in the input space is mapped to a convex tile $M(\omega)$ on the manifold using the affine transform 
\begin{equation} 
    M(\omega) = \{\bA_{\omega} \bx + \bc_{\omega}, \: \bx \in \omega \},
    \label{eq:region_mapping}
\end{equation}
and the manifold ${\cal M}$ is the union of the $M(\omega)$.
This straightforward construction enables us to analytically characterize many properties of ${\cal M}$ via (\ref{eq:spline1}).

In particular, it is easy to show that the mapping (\ref{eq:region_mapping})  from the input space to the manifold {\em shears} the tiles in the input space tessellation by $\bA_\omega$, causing their volume to expand or contract by the factor
\begin{equation}
    \frac{{\sf vol}(M(\omega))}{{\sf vol}(\omega)} = \sqrt{\det(\bA_{\omega}^\top \bA_{\omega})}.
    \label{eq:vol}
\end{equation}
Knowing this, we can take any trained and fixed generative model and determine a {\em nonuniform sampling} of the input space according to (\ref{eq:vol}) such that the sampling on the manifold is provably uniform and free from bias.
The bonus is that this procedure, which we call MAximum entropy Generative NETwork (MaGNET) \cite{magnet}, is simply a post-processing procedure that does not require any retraining of the network.

\begin{figure}[h]
    \centering
    \begin{minipage}{0.45\linewidth}
    \centering
    \includegraphics[width=\linewidth]{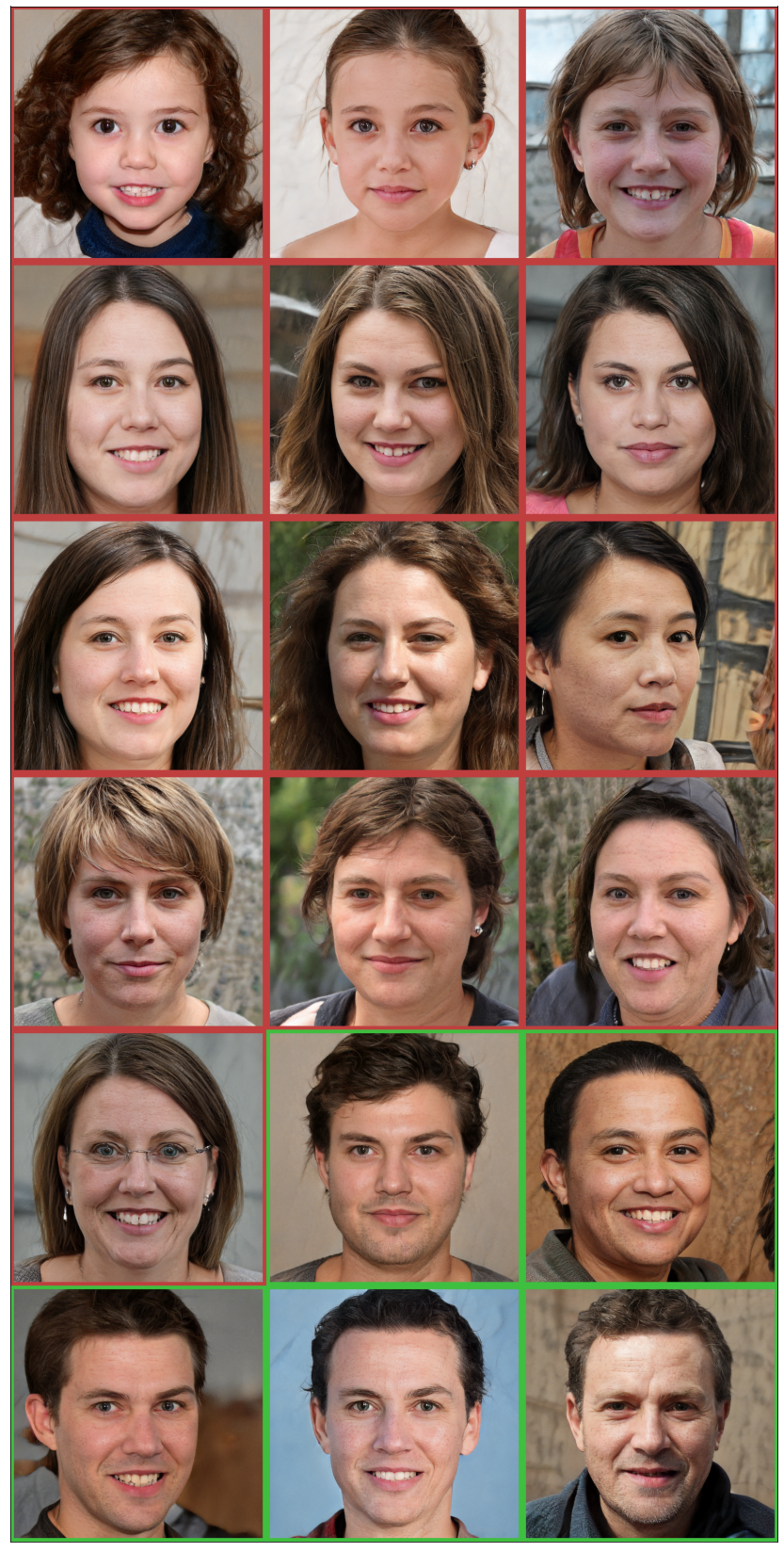}
    \\ {\small\sf StyleGAN2}
    \end{minipage}
    \hspace*{2mm}
    \begin{minipage}{0.45\linewidth}
    \centering
    \includegraphics[width=\linewidth]{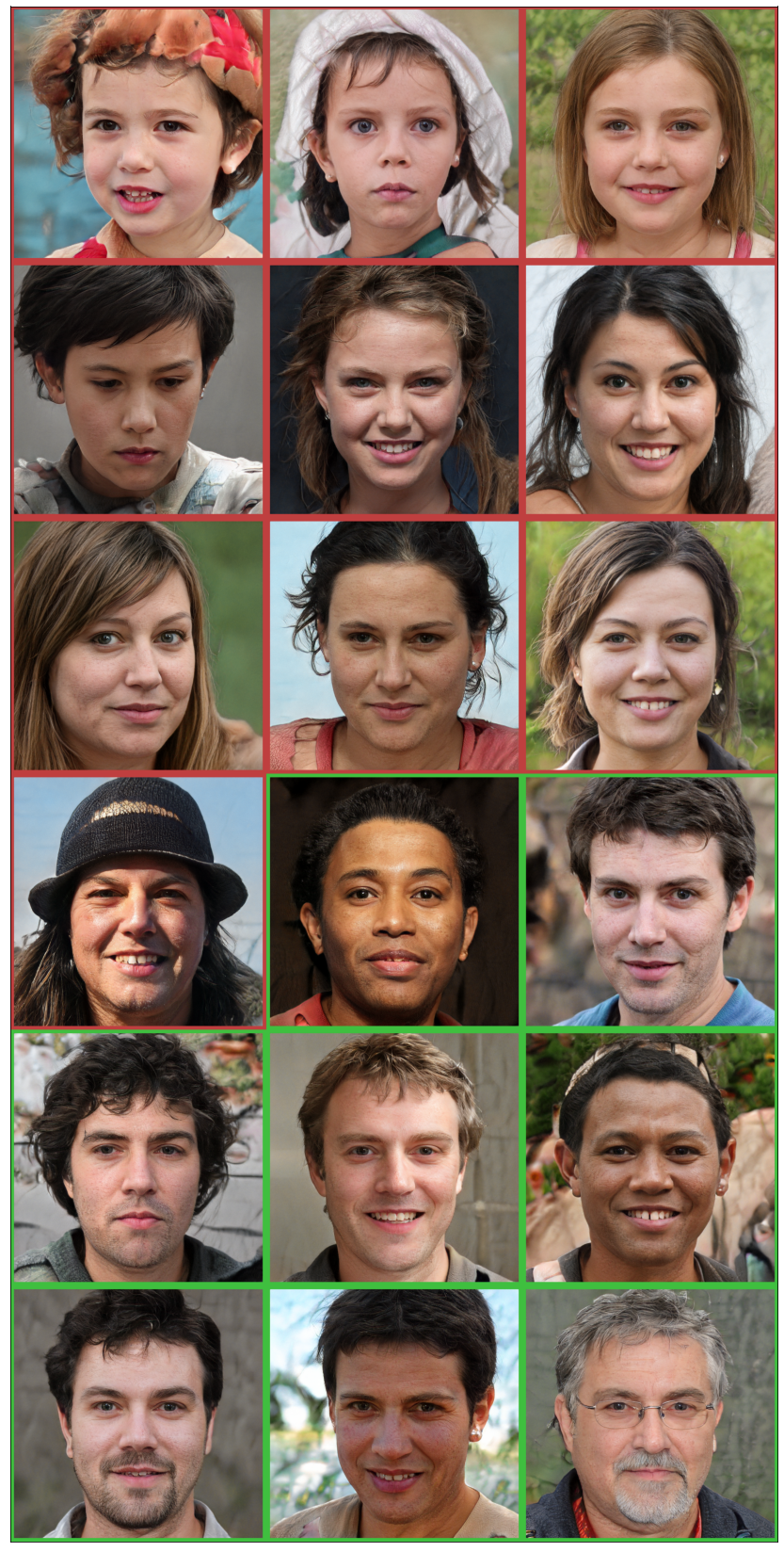}
    \\ {\small\sf MaGNET-StyleGAN2}
    \end{minipage}
    \caption{\small
    Images synthesized by sampling uniformly from the input space of a StyleGAN2 deep generative model trained on the FFHQ face data set and nonuniformly according to (\ref{eq:vol}) using MaGNET.
    (Adapted from \cite{magnet}.)
}
    \label{fig:magnet}
\end{figure}

Figure~\ref{fig:magnet} demonstrates MaGNET's debiasing abilities.
On the left are 18 faces synthesized by the StyleGAN2 generative model trained on the FFHQ face dataset.
On the right are 18 faces synthesized by the same StyleGAN2 generative model but using a nonuniform sampling distribution on the input space based on (\ref{eq:vol}).
MaGNET sampling yields a better gender, age, and hair color balance as well as more diverse backgrounds and accessories.
In fact, MaGNET sampling produces 41\% more male faces (as determined by the Microsoft Cognitive API) to balance out the gender distribution.

We can turn the volumetric deformation (\ref{eq:vol}) into a tool to efficiently explore the data distribution on a deep generative model's manifold.
By following the MaGNET sampling approach but using an input sampling distribution based on 
$\det(\bA_{\omega}^\top \bA_{\omega})^\rho$ we can synthesize images in the {\em modes} (high probability regions of the manifold 
that are more ``typical and high quality'') using ${\rho} \to -\infty$ and or in the {\em anti-modes} (low probability regions of the manifold that are more ``diverse and exploratory'') using ${\rho} \to \infty$ \cite{humayun2022polarity}. 
Setting $\rho=0$ returns the model to uniform sampling.

\begin{figure}[h]
    \centering  
\includegraphics[width=1\linewidth]{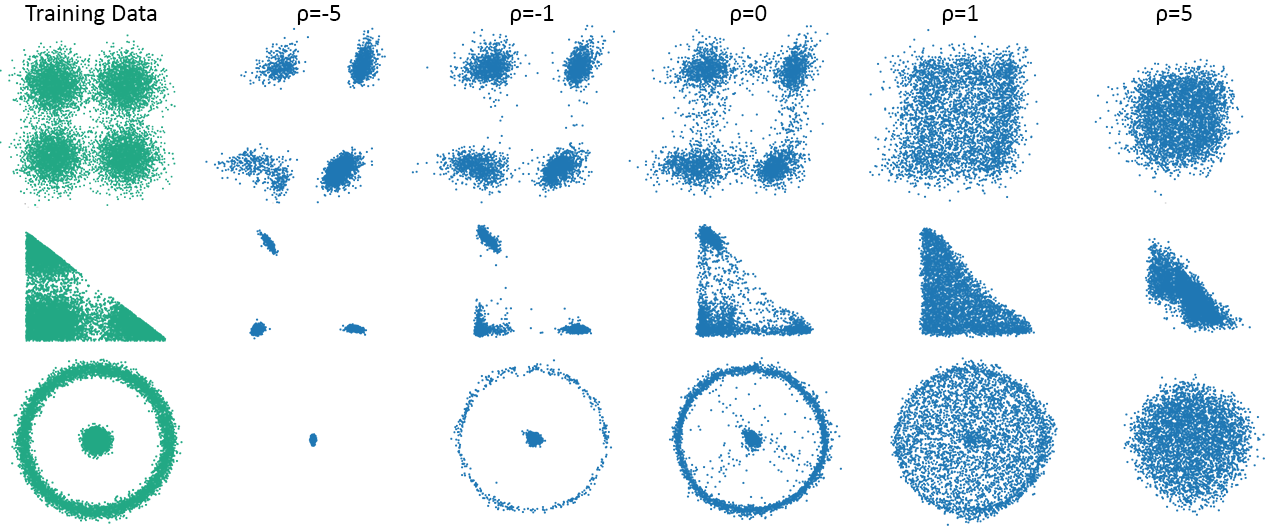}
\vspace*{-5mm}
    \caption{\small
    Polarity-guided synthesis of points in the plane by a Wasserstein GAN generative model.
    When the polarity parameter $\rho=0$, the model produces a data distribution closely resembling the training data.
    When the polarity parameter $\rho\ll 0$ ($\rho\gg 0$), the WGAN produces a data distribution focusing on the modes (anti-modes), the high (low) probability regions of the training data.
    (From \cite{humayun2022polarity}.)}
    \label{fig:polarity}
\end{figure}

Like MaGNET, this {\em polarity sampling} approach applies to any pre-trained generative network and so has broad applicability.
See Figure~\ref{fig:polarity} for an illustrative toy example and \cite{humayun2022polarity} for numerous examples with large-scale generative models, including using polarity sampling to boost the performance of existing generative models to state-of-the-art.

There are many interesting open research questions around affine splines and deep generative networks. One related to the  MaGNET sampling strategy is that it assumes that the trained generative network actually learned a good enough approximation of the true underlying data manifold. 
One could envision exploring how MaGNET could be used to test such an assumption. 

\subsection*{Discussion and Outlook}

While there are several ways to envision extending the concept of a one-dimensional affine spline (recall Figure~\ref{fig:1Dspline}) to high-dimensional functions and operators, progress has been made only along the direction of forcing the tessellation of the domain to hew to some kind of grid (e.g., uniform or multiscale uniform for spline wavelets).
Such constructions are ill-suited for machine learning problems in high dimensions due to the so-called curse of dimensionality that renders approximation intractable.

We can view deep networks as a tractable mechanism for emulating those most powerful of splines, the free-knot splines (splines like those in Figure~\ref{fig:1Dspline} where the intervals partitioning the real line domain are arbitrary) in high dimensions.
A deep network uses the power of a hyperplane arrangement to tractably create a myriad of flexible convex polytopal tiles that tessellate its input space plus affine transformations on each that result in quite powerful approximation capabilities in theory \cite{devore2021neural} and in practice.
There is much work to do in studying these approximations (e.g., developing realistic function approximation classes and proving approximation rates) as well as developing new deep network architectures that attain improved rates and robustness.

An additional timely research direction involves extending the ideas discussed here to deep networks like {\em transformers} that employ at least some nonlinearities that are not piecewise linear.
The promising news is that the bulk of the learnable parameters in state-of-the-art transformers lie in readily analyzable affine spline layers within each transformer block of the network.
Hence, we can apply many of the above ideas, including local complexity (LC) estimation, to study the smoothness, expressivity, and sensitivity characteristics of even monstrously large language models like the GPT, Gemini, and Llama series.

We hope that we have convinced you that viewing deep networks as affine splines provides a powerful geomeric toolbox to better understand how they learn, how they predict, and how they can be improved in a principled fashion.
But splines are just one interesting research direction in the mathematics of deep learning.
These are early days, and there are many more open than closed research questions.

\subsection*{Acknowledgments}


Thanks to T.\ Mitchell Roddenberry and Ali Siahkoohi for their comments on the manuscript.
AIH and RGB were supported by NSF grants CCF-1911094, IIS-1838177, and IIS-1730574; ONR grants N00014-18-1-2571, N00014-20-1-2534, N00014-23-1-2714, and MURI N00014-20-1-2787; AFOSR grant FA9550-22-1-0060; DOI grant 140D0423C0076; and a Vannevar Bush Faculty Fellowship, ONR grant N00014-18-1-2047.

\bibliography{ref}
\end{document}